\def\year{2021}\relax
\documentclass[letterpaper]{article} 
\usepackage{aaai21}  
\usepackage{times}  
\usepackage{helvet} 
\usepackage{courier}  
\usepackage[hyphens]{url}  
\usepackage{graphicx} 
\usepackage{natbib}  
\usepackage{caption} 
\nocopyright
\usepackage{amsmath}
\usepackage{amssymb}
\usepackage[switch]{lineno}
\title{TSQA: Tabular Scenario Based Question Answering}
\author {
        Xiao Li,
        Yawei Sun,
        Gong Cheng \\
}
\affiliations {
    State Key Laboratory for Novel Software Technology, Nanjing University, China \\
    \{xiaoli.nju, ywsun\}@smail.nju.edu.cn, gcheng@nju.edu.cn
}

\def \DATASETNAME {GeoTSQA}
\def \MODELNAME {TTGen}

\usepackage{xspace}
\newcommand{\softmax}{\ensuremath{\mathtt{softmax}}\xspace}
\newcommand{\sigmoid}{\ensuremath{\mathtt{sigmoid}}\xspace}

\begin{document}

\maketitle

\begin{abstract}
Scenario-based question answering (SQA) has attracted an increasing research interest. Compared with the well-studied machine reading comprehension (MRC), SQA is a more challenging task: a scenario may contain not only a textual passage to read but also structured data like tables, i.e.,~tabular scenario based question answering (TSQA). AI applications of TSQA such as answering multiple-choice questions in high-school exams require synthesizing data in multiple cells and combining tables with texts and domain knowledge to infer answers. To support the study of this task, we construct \DATASETNAME. This dataset contains 1k~real questions contextualized by tabular scenarios in the geography domain. To solve the task, we extend state-of-the-art MRC methods with \MODELNAME, a novel table-to-text generator. It generates sentences from variously synthesized tabular data and feeds the downstream MRC method with the most useful sentences. Its sentence ranking model fuses the information in the scenario, question, and domain knowledge. Our approach outperforms a variety of strong baseline methods on \DATASETNAME.
\end{abstract} 

\section{Introduction}

Scenario-based question answering (SQA) is to answer questions contextualized by scenarios~\cite{lally2017watsonpaths}.
Compared with the well-studied task of machine reading comprehension~(MRC) which requires reading a passage to extract or infer an answer~\cite{rajpurkar-etal-2016-squad,lai-etal-2017-race}, a SQA task requires reading a scenario which commonly contains both a textual passage and a set of structured data. One such prominent AI application of SQA is answering multiple-choice questions in high-school geography exams~\cite{ding2018answering,huang-etal-2019-geosqa}. Those questions are contextualized by scenarios containing tables and diagrams, where the rich information cannot be captured by current MRC methods but have to be manually interpreted using natural language. Thus, one natural research question arises: can we solve SQA in a fully automated manner?

\begin{figure*}[t]
    \centering
    \includegraphics[width=0.9\textwidth]{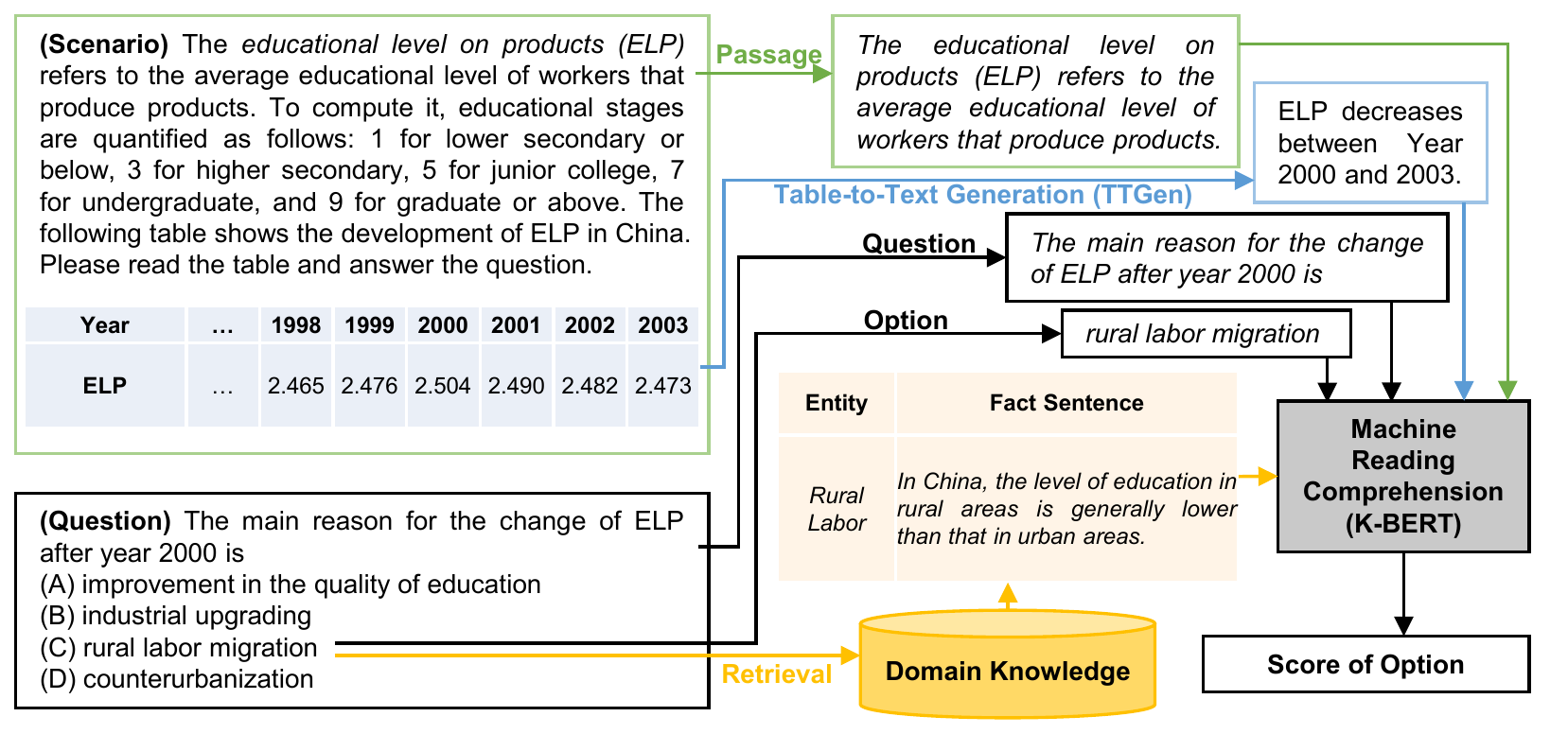}
    \caption{Left: an example question contextualized by a tabular scenario in \DATASETNAME. Right: an overview of our approach.}
    \label{fig:sample}
\end{figure*}

\subsubsection{Task and Challenges.}

Specifically, we focus on questions contextualized by a scenario consisting of a textual passage and a set of tables. We refer to this branch of SQA as \textbf{TSQA}, short for \emph{Tabular Scenario based Question Answering}. To support the study of this task, we construct a dataset named \textbf{\DATASETNAME}. It contains 1k~real questions contextualized by tabular scenarios in the geography domain, collected from China's high-school exams. Compared with existing datasets for table-based question answering like WikiTableQuestions~\cite{pasupat-liang-2015-compositional}, \DATASETNAME\ requires fundamentally different reading and reasoning skills, and poses new research challenges.

For instance, Figure~\ref{fig:sample} shows a question in \DATASETNAME. To answer it, tabular data needs to be synthesized via a complex operation: identifying a monotonic increase in ELP over the interval 2000--2003. Focusing on this particular interval rather than many other intervals is implicitly suggested in the question: after year~2000. Moreover, the passage in the scenario helps to link ELP with educational level, and the retrieved domain knowledge bridges the gap between educational level and rural labor which is the correct answer. To conclude, TSQA methods need to \emph{properly manipulate tabular data}, and \emph{comprehend fused textual information}.

\subsubsection{Our Approach.}

To meet the challenges, considering that text reading has been extensively studied in MRC research, we propose to extend state-of-the-art MRC methods with a novel table-to-text generator named \textbf{\MODELNAME} to specifically handle tabular data. The basic idea is straightforward: feeding a MRC model with sentences generated from tables \emph{using templates that encapsulate many and various predefined operations for manipulating tabular data}. However, the potentially large number (e.g.,~hundreds) of generated sentences may easily exceed the capacity of typical MRC models, and produce much noise information influencing the accuracy of reading comprehension. To address this problem, \MODELNAME\ incorporates a sentence ranking model that fuses the information in the scenario, question, and domain knowledge to effectively \emph{select sentences that are most useful for answering the question}. It outperforms a variety of strong baseline methods in extensive experiments on \DATASETNAME.

We summarize our contributions in the paper as follows.
\begin{itemize}
    \item We construct and publish \DATASETNAME, the first dataset dedicated to TSQA. It requires reading and reasoning with tables, texts, and domain knowledge at high school level.
    \item We extend MRC methods with TTGen to solve TSQA. TTGen performs question and knowledge aware ranking of sentences generated from synthesized tabular data.
\end{itemize}

\subsubsection{Outline.}

The remainder of the paper is organized as follows. We discuss and compare with related work in Section~\ref{sec:relatedwork}. We formally define the TSQA task and describe the construction of the \DATASETNAME\ dataset in Section~\ref{sec:task-dataset}. We introduce our approach in Section~\ref{sec:approach}. We present experiment settings in Section~\ref{sec:exp-setup} and report experiment results in Section~\ref{sec:exp-result}. Finally we conclude the paper in Section~\ref{sec:conclusion}.

Our code and data are available on Github.\footnote{https://github.com/nju-websoft/TSQA}

\section{Related Work}
\label{sec:relatedwork}

\subsection{SQA}

SQA is an emerging AI task and has found application in many domains. The pioneering WatsonPaths system provides recommendations for diagnosis and treatment based on a medical scenario about a patient~\cite{lally2017watsonpaths}. In the legal domain, SQA supports judgment prediction based on the fact description of a legal case~\cite{ye-etal-2018-interpretable,zhong2018legal,DBLP:conf/ijcai/YangJZL19}.

We focus on TSQA where a scenario contains both textual and tabular data. Such questions are common in, for example, China's high-school geography and history exams where a scenario describes a concrete fact or event to contextualize a set of questions. Previous efforts in this domain either ignore tables~\cite{DBLP:conf/ijcai/ChengZWCQ16} or manually transform tables into triple-structured knowledge~\cite{ding2018answering} or natural language descriptions for machine reading~\cite{huang-etal-2019-geosqa}. In contrast, we aim at \emph{solving TSQA in a fully automated manner by generating texts from tables}.

\subsection{Table-to-Text Generation}

Table-to-text generation has been studied for decades. Early methods rely on handcrafted rules to generate texts for specific domains such as stock market summaries~\cite{kukich-1983-design} and weather forecasts~\cite{294135}.
They typically implement a pipeline of modules including content planning, sentence planning, and surface realization. Today, it is feasible to train neural generation models in an end-to-end fashion, thanks to the availability of effective pre-trained language models~\cite{devlin-etal-2019-bert,radford2019language} and large datasets~\cite{lebret-etal-2016-neural,wiseman-etal-2017-challenges,Dusek2019EvaluatingTS}. Current models often adopt an encoder-decoder architecture with a copy mechanism~\cite{wiseman-etal-2017-challenges,Puduppully2019DatatoTextGW}. Moreover, they can be enhanced with entity representations~\cite{puduppully-etal-2019-data} and external background knowledge~\cite{chen-etal-2019-enhancing}.

The above methods are targeted on surface-level description of tabular data, which is insufficient for our task where data in multiple cells needs to be \emph{synthesized using various operations} (e.g.,~extremum, monotonicity, trend). Generating such natural language statements that are logically entailed from tabular data, rather than superficial restatements, has recently attracted research attention~\cite{Chen2020,Chen2020Logic2TextHN}.
However, they are primarily focused on high-fidelity generation, i.e.,~the generated text should be faithful to the tabular data. Fidelity is necessary but insufficient for our task where the generated text also needs to be useful for answering the question. It is thus essential to \emph{select the proper operation and data from a potentially very large space}. To this end, our proposed generator \MODELNAME\ features a sentence ranking model that fuses the information in the scenario, question, and domain knowledge.

\subsection{Table-Based Question Answering}

Similar to TSQA, there has been a line of research of answering questions over tabular data~\cite{pasupat-liang-2015-compositional,jauhar2016tables,DBLP:conf/ijcai/YinLLK16,yu2020dataset}. Like our constructed dataset \DATASETNAME, these datasets also require performing various operations over multiple cells. Differently, their questions can be answered solely on the basis of tabular data, whereas the questions in \DATASETNAME\ are more naturally contextualized by a scenario containing \emph{both} a set of tables and a textual passage which are equally important and are \emph{dependent on each other}.

From this angle, the most similar dataset to \DATASETNAME\ is HybridQA~\cite{Chen2020HybridQAAD}, where table cells are linked with Wikipedia pages. However, \DATASETNAME\ has its \emph{unique challenges} due to the source of questions---high-school geography exams. For example, table cells mainly contain non-linkable numeric values; more complex operations (e.g.,~monotonicity) are needed; it would be helpful to incorporate domain knowledge into question answering.

\section{Task and Dataset}
\label{sec:task-dataset}

We firstly define the task of TSQA, and then we construct the \DATASETNAME\ dataset to support the study of TSQA.

\subsection{Task Definition}

A TSQA task consists of a scenario $\langle P,T \rangle$, a question~$Q$, and a set of options~$O$ as candidate answers of which only one is correct. The scenario contains a passage~$P$ and a set of tables~$T$. Each table in~$T$ has a header row, a header column, and a set of content cells.
The goal is to select an option from~$O$ as the answer to~$Q$ contextualized by~$\langle P,T \rangle$.

\subsection{Dataset Construction}
\label{sec:dataset}

We constructed \DATASETNAME. To the best of our knowledge, it is the first dataset dedicated to the TSQA task.


\subsubsection{Collecting Questions.}
We collected multiple-choice questions contextualized by tabular scenarios in the geography domain from China's high-school exams. A related dataset is GeoSQA~\cite{huang-etal-2019-geosqa}. We not only collected all the questions from GeoSQA but also reused the code for constructing GeoSQA to crawl much more questions from the Web to expand our dataset.

However, many collected scenarios are not tabular. Indeed, each scenario is associated with a set of image files. Each image file depicts either a table or another kind of diagram such as a map or a histogram. Therefore, we need to identify images depicting tables or table-like diagrams.

\subsubsection{Identifying Tables.}

We looked for tables, or charts that can be straightforwardly converted to tables (e.g.,~histograms, line charts). We manually identified 200~such image files as positive examples and another 200~image files as negative examples. We used them to train an image classifier~\cite{DBLP:conf/cvpr/SzegedyVISW16} to classify all the remaining image files. Finally, for all the image files that were classified as positive, we manually checked them for classification errors.

\subsubsection{Extracting Tables.}

We recruited 15~undergraduate students from a university in China as annotators. For image files depicting tables, we used Baidu's OCR tool to extract tabular data. OCR errors were manually corrected by annotators. For image files depicting charts, annotators manually extracted tabular data, assisted with a tool we developed. The annotator used that tool to easily click key points in the image, e.g.,~the origin, coordinate axes, data points. The tool then automatically converted data points to data tables.

Annotators manually checked each extracted table and filtered out irregular tables (e.g.,~with multi-level headers).

\subsubsection{Filtering Questions.}

Last but not least, annotators filtered out questions that can be answered without using any table. Therefore, every question in \DATASETNAME\ is contextualized by a tabular scenario, and it is essential to employ the information in the given tables to answer the question.

\subsection{Dataset Statistics}

\begin{table}[t!]
\centering
\small
\begin{tabular}{|l|rl|}
    \hline
    Scenarios & 556 & \\
    Chinese characters per passage & $52.42$ & $\pm 32.99$ \\
    Tables per scenario & $1.58$ & $\pm 0.93$ \\
    Cells per table & $26.98$ & $\pm 17.51$ \\
    \hline
    Questions & 1,012 & \\
    Chinese characters per question & $44.02$ & $\pm 15.89$ \\
    \hline
\end{tabular}
\caption{Statistics about \DATASETNAME.}
\label{table:datasetstatistics}
\end{table}

\DATASETNAME\ contains 556~scenarios and 1,012~multiple-choice questions. Each question has four options. More statistics about the dataset are shown in Table~\ref{table:datasetstatistics}.

Out of the 878~tables in \DATASETNAME, 96\%~only contain numeric content cells.
It differs from HybridQA~\cite{Chen2020HybridQAAD} where content cells are often entities linked with Wikipedia pages, thereby providing extra background knowledge for answering questions. For \DATASETNAME, to obtain information that is not explicitly given in the scenario but critical for answering questions, it is essential to entail from tabular data via operations over multiple cells.

\section{Approach}
\label{sec:approach}

We propose a two-step approach to solve TSQA. As illustrated in Figure~\ref{fig:sample}, the first step~(Section~\ref{sec:approach-ttgen}) is a table-to-text generator named \MODELNAME. From the tables~$T$ in a scenario~$\langle P,T \rangle$, \MODELNAME\ generates top-$k$ sentences~$S$ that are most useful for answering the question~$Q$. The second step~(Section~\ref{sec:approach-mrc}) is a MRC method based on K-BERT~\cite{Liu2019KBERTEL}, a state-of-the-art knowledge-enabled language model. It fuses the information in the passage~$P$, generated sentences~$S$, question~$Q$, and domain knowledge~$K$ to rank the options in~$O$.

\subsection{MRC with Domain Knowledge}
\label{sec:approach-mrc}

Our MRC method is based on K-BERT~\cite{Liu2019KBERTEL}. This state-of-the-art language model extends BERT~\cite{devlin-etal-2019-bert} with the capability to utilize external knowledge such as domain knowledge.

\subsubsection{MRC with K-BERT.}

For each option $o_i \in O$, we concatenate the passage~$P$, top-$k$ sentences $S=\{s_1,\ldots,s_k\}$ generated from the tables~$T$, question~$Q$, and~$o_i$ in a standard way, starting with a [CLS] token and separating with [SEP]:
\begin{equation}
\resizebox{\columnwidth}{!}{$
    I^\text{MRC}_i = \text{[CLS] $P$ $s_1 \cdots s_k$ $Q$ [SEP] $o_i$ [SEP] \emph{NUMS}$_i$ [SEP]} \,,
$}
\end{equation}
\noindent where \emph{NUMS}$_i$ is a concatenation of all the numeric tokens in~$P$, $S$, $Q$, and~$o_i$. Each numeric token in the original position is replaced by a special token [NUM].

We use K-BERT to obtain a vector representation for each token in~$I^\text{MRC}_i$ to capture its semantic features:
\begin{equation}
    \langle \mathbf{h}^\text{MRC}_{i1}, \mathbf{h}^\text{MRC}_{i2}, \ldots \rangle = \text{K-BERT}(I^\text{MRC}_i, ~K) \,,
\end{equation}
\noindent where $K$~is an external knowledge base we will explain later.

The vector representation for the [CLS] token, i.e.,~$\mathbf{h}^\text{MRC}_{i1}$, is used as an aggregate representation for~$I^\text{MRC}_i$. It is fed into two dense layers followed by a softmax layer to obtain a correctness score~$\hat{\omega}_i$ for each option $o_i \in O$:
\begin{equation}
\begin{split}
    \omega_i & = \mathbf{w}_2^\intercal \tanh(\mathbf{W}_1 \mathbf{h}^\text{MRC}_{i1} + \mathbf{b}_1) + b_2 \,,\\
    \mathbf{\Omega} & =[\hat{\omega}_1; \hat{\omega}_2; \ldots] = \softmax([\omega_1; \omega_2; \ldots]) \,,
\end{split}
\end{equation}
\noindent where $\mathbf{W}_1$~is a trainable matrix, $\mathbf{w}_2$ and~$\mathbf{b}_1$ are trainable vectors, and $b_2$~is a trainable parameter.

In the training phase, we minimize the negative log-likelihood loss which measures the difference between~$\mathbf{\Omega}$ and the binary correctness label on each option (we will detail in Section~\ref{sec:exp-label}). In the test phase, we choose the option in~$O$ with the highest correctness score~$\hat{\omega}$ as the answer.

K-BERT extends BERT with an external knowledge base~$K$. It helps to fuse the information in~$P$, $S$, $Q$, $O$, and~$K$. We refer the reader to \citet{Liu2019KBERTEL} for a detailed description of K-BERT. Briefly, each entry in~$K$ is a pair $\langle \text{entity}, ~\text{fact sentence}\rangle$, or a triple $\langle \text{entity}, ~\text{property}, ~\text{value} \rangle$ which can be converted into a pair by concatenating the property and the value into a fact sentence. K-BERT employs~$K$ to expand the input sequence into a tree of tokens: fact sentences about an entity are retrieved from~$K$ and inserted as branches after each mention of the entity in the input sequence.
In our implementation, for each entity, we retrieve top-$\epsilon$~fact sentences that are most relevant to the input sequence. The relevance of a fact sentence to the input sequence is measured by the cosine similarity between their average pre-trained BERT embedding vectors.

\subsubsection{Domain Knowledge.}

For the external knowledge base~$K$, for our experiments we use domain knowledge since all the questions in \DATASETNAME\ are in the geography domain. We obtain domain knowledge from two sources.

First, we import all the triples in Clinga~\cite{10.1007/978-3-319-46547-0_11}, a large Chinese geographical knowledge base.

Second, we reuse the corpus in~\cite{huang-etal-2019-geosqa}. The corpus contains a geography textbook providing a set of entity descriptions. We pair each entity with each sentence in its description as a fact sentence. The corpus also contains a subset of Chinese Wikipedia. We treat the title of each page as an entity and pair it with each sentence in the page as a fact sentence.

\subsection{Table-to-Text Generation (\MODELNAME)}
\label{sec:approach-ttgen}

Below we describe the generation of sentences from tables to be fed into our MRC method. We rely on templates that encapsulate predefined operations for manipulating tabular data. It enables us to perform complex operations that are needed for answering hard questions such as those in \DATASETNAME. We generate sentences from tables using all the applicable templates. However, it is infeasible for a MRC model like K-BERT to jointly encode a large number (e.g.,~hundreds) of sentences. Therefore, we rank the generated sentences and select $k$~top-ranked sentences that are most useful for answering the question. By filtering the generated sentences, we can also reduce noise information that may influence the accuracy of reading comprehension.

\subsubsection{Sentence Generation.}

By significantly extending the operations considered in~\citet{Chen2020,DBLP:conf/iclr/ChenWCZWLZW20}, we define six table-to-text templates that encapsulate different powerful operations for synthesizing numeric tabular data. As we will show in the experiments, these templates have covered most needs about tables in \DATASETNAME. One can easily add new templates to accommodate other applications.
\begin{itemize}
    \item \textbf{Extremum.} This template reports the maximum or minimum value of a row or column. An example sentence generated from the table in Figure~\ref{fig:sample} is: \emph{ELP reaches a maximum of~2.504 at Year~2000.}
    \item \textbf{Special values.} This template reports or compares with a special value (e.g.,~under a column header that is mentioned in the question), e.g.,~\emph{ELP at Year~2000 is~2.504.}
    \item \textbf{Comparison with average.} This template reports a maximal sequence of cells where all the values are above or below the average of the entire row or column, e.g.,~\emph{ELP is relatively large between Year~2000 and~2002.}
    \item \textbf{Monotonicity.} This template reports a monotonic increase or decrease over a maximal sequence of cells, e.g.,~\emph{ELP decreases between Year~2000 and~2003.}
    \item \textbf{Trend.} This template reports the overall trend of a row or column, e.g.,~\emph{ELP generally increases and then decreases.}
    \item \textbf{Range comparison.} This template reports a comparison between two maximal corresponding sequences of cells from different rows or columns.
\end{itemize}

For non-numeric tabular data, we simply concatenate each row header, each column header, and the corresponding content cell into a sentence.

\subsubsection{Sentence Ranking.}

\begin{figure}[t]
    \centering
    \includegraphics[width=1.0\columnwidth]{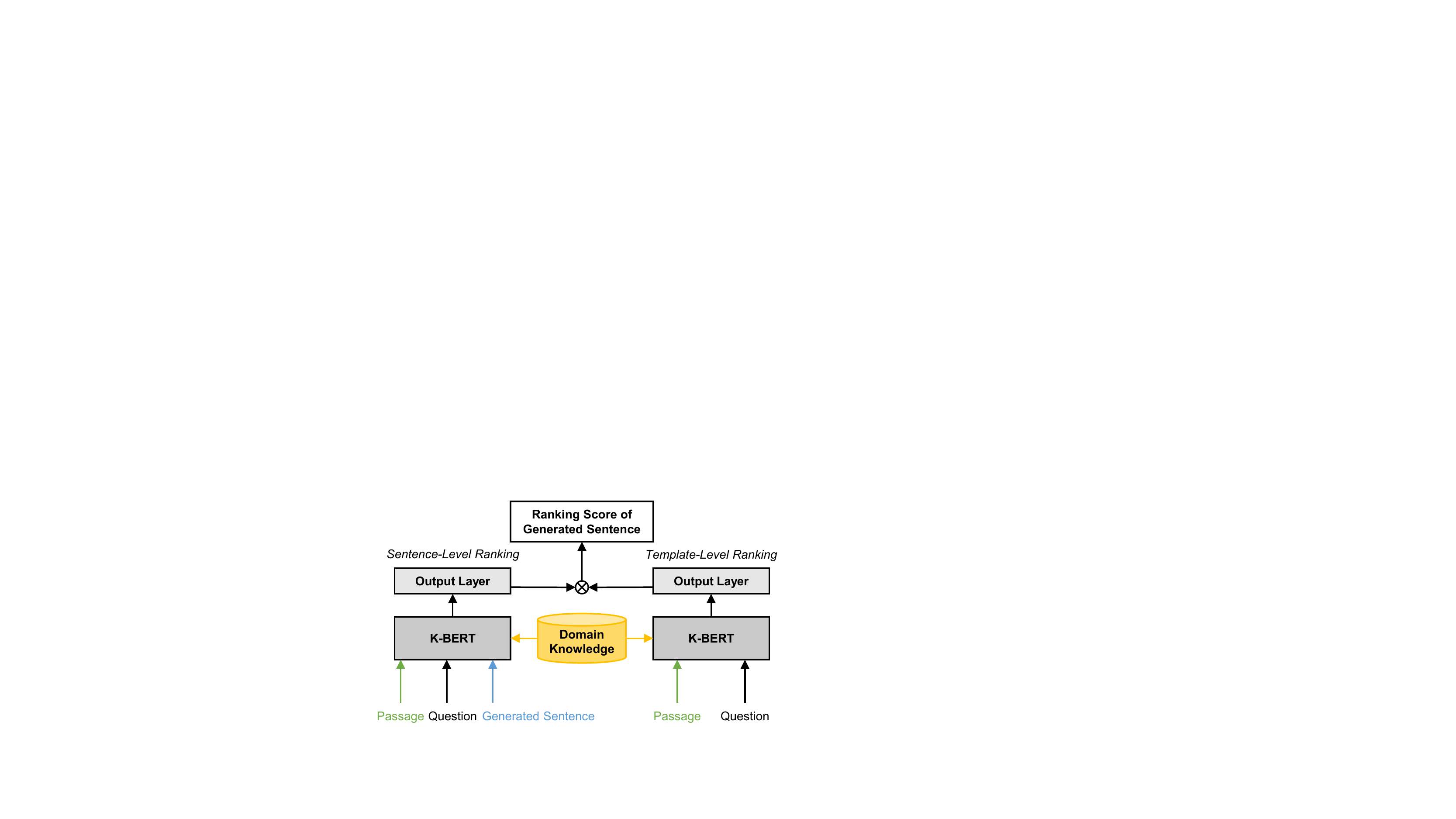}
    \caption{Sentence ranking model in \MODELNAME.}
    \label{fig:model}
\end{figure}

Let~$\hat{S}$ be the set of sentences generated from the tables~$T$ using all the applicable templates. We compute a usefulness score for each sentence $s_j \in \hat{S}$, and choose $k$~top-ranked sentences $S \subseteq \hat{S}$. To select sentences that are most useful for answering the question, our ranking model employs K-BERT to fuse the information in the passage~$P$, question~$Q$, and domain knowledge~$K$ to perform question and knowledge aware ranking. Figure~\ref{fig:model} presents an overview of the model. It integrates two complementary rankers: sentence-level ranking directly assesses the usefulness of each individual sentence; template-level ranking infers useful templates purely from the passage and question.

For sentence-level ranking, we concatenate the passage~$P$, question~$Q$, and sentence~$s_j$ in a standard way:
\begin{equation}
    I^\text{SR}_j = \text{[CLS] $P$ $Q$ [SEP] $s_j$ [SEP] \emph{NUMS}$_j$ [SEP]} \,,
\end{equation}
\noindent where \emph{NUMS}$_j$ is a concatenation of all the numeric tokens in~$P$, $Q$, and~$s_j$. Each numeric token in the original position is replaced by a special token [NUM]. We use K-BERT to obtain a vector representation for each token in~$I^\text{SR}_j$:
\begin{equation}
    \langle \mathbf{h}^\text{SR}_{j1}, \mathbf{h}^\text{SR}_{j2}, \ldots \rangle = \text{K-BERT}(I^\text{SR}_j, ~K) \,.
\end{equation}
\noindent The vector representation for the [CLS] token, i.e.,~$\mathbf{h}^\text{SR}_{j1}$, is fed into two dense layers followed by a softmax layer to obtain a usefulness score~$\hat{\phi}_j$ for each sentence $s_j \in \hat{S}$:
\begin{equation}
\begin{split}
    \phi_j & = \mathbf{w}_4^\intercal \tanh(\mathbf{W}_3 \mathbf{h}^\text{SR}_{j1} + \mathbf{b}_3) + b_4 \,,\\
    \mathbf{\Phi} & = [\hat{\phi}_1; \hat{\phi}_2; \ldots] = \softmax([\phi_1; \phi_2; \ldots]) \,,
\end{split}
\end{equation}
\noindent where $\mathbf{W}_3$~is a trainable matrix, $\mathbf{w}_4$ and~$\mathbf{b}_3$ are trainable vectors, and $b_4$~is a trainable parameter. In the training phase, we minimize the negative log-likelihood loss which measures the difference between~$\mathbf{\Phi}$ and the binary usefulness label on each generated sentence (we will detail in Section~\ref{sec:exp-label}).

For template-level ranking, we concatenate the passage~$P$ and question~$Q$ in a standard way:
\begin{equation}
    I^\text{TR} = \text{[CLS] $P$ $Q$ [SEP]} \,.
\end{equation}
\noindent We use K-BERT to obtain a vector representation for each token in~$I^\text{TR}$:
\begin{equation}
    \langle \mathbf{h}^\text{TR}_{1}, \mathbf{h}^\text{TR}_{2}, \ldots \rangle = \text{K-BERT}(I^\text{TR}, ~K) \,.
\end{equation}
\noindent The vector representation for the [CLS] token, i.e.,~$\mathbf{h}^\text{TR}_{1}$, is fed into two dense layers followed by a sigmoid layer to obtain a usefulness score~$\hat{\psi}$ for each of the six templates:
\begin{equation}
\begin{split}
    [\psi_1; \ldots; \psi_6] & = \mathbf{W}_6 \tanh(\mathbf{W}_5 \mathbf{h}^\text{TR}_1 + \mathbf{b}_5) + \mathbf{b}_6 \,,\\
    \mathbf{\Psi} & = [\hat{\psi}_1; \ldots; \hat{\psi}_6] = \sigmoid([\psi_1; \ldots; \psi_6]) \,,
\end{split}
\end{equation}
\noindent where $\mathbf{W}_5$ and~$\mathbf{W}_6$~are trainable matrices, $\mathbf{b}_5$ and~$\mathbf{b}_6$ are trainable vectors. Let sentence~$s_j$ be generated by the $\tau_j$-th template. We derive usefulness labels on templates for training from usefulness labels on generated sentences: a template is labeled useful if and only if at least one sentence it generates is labeled useful. Multiple sentences and hence multiple templates may be labeled useful for answering a question. Therefore, in the training phase, we formulate a multi-label binary classification task, and we minimize the binary cross-entropy loss which measures the difference between~$\mathbf{\Psi}$ and the binary usefulness label on each template.

Finally, in the test phase, we compute:
\begin{equation}
    \text{usefulness score of } s_j = \hat{\phi}_j \cdot \hat{\psi}_{\tau_j} \,.
\end{equation}



\section{Experiment Setup}
\label{sec:exp-setup}

We compared our approach with a variety of strong baseline methods for TSQA. We also evaluated our sentence ranking model, which is the core component of our approach.

\subsection{Labeled Data}
\label{sec:exp-label}


\subsubsection{Correctness Labels on Options.}

For each question, from its known correct answer, we derived a label for each of the four options indicating whether it is the correct answer. These binary correctness labels were used to train and evaluate TSQA methods.

\subsubsection{Usefulness Labels on Generated Sentences.}

The number of all the sentences~$\hat{S}$ generated by our templates for a question is in the range of 2--176, with a mean of~41.58 and a median of~38. For each question, we asked an annotator (recruited in Section~\ref{sec:dataset}) to read~$\hat{S}$ and assign a label to each sentence indicating whether it is useful for answering the question. These binary usefulness labels were used to train and evaluate sentence ranking models.

\subsubsection{Gold-Standard Sentences.}

Furthermore, the annotator manually summarized the tables in one sentence describing necessary information for answering the question. This gold-standard sentence was used for comparison.

We randomly sampled 100~questions from \DATASETNAME. For 92~questions, $\hat{S}$~fully covers the information in the gold-standard sentence. For 6~questions, $\hat{S}$~partially covers that information. Therefore, our six templates show good coverage of the various operations required by \DATASETNAME.

\subsection{Baselines}

Our approach extends MRC methods. It is not our focus to compare existing MRC methods. Instead, table-to-text generation is our major technical contribution. Therefore, in the experiments we consistently used the MRC method based on K-BERT described in Section~\ref{sec:approach-mrc}, but fed it with sentences generated from tables by the following different methods.

\subsubsection{Supervised Methods.}

\begin{table}[t!]
\centering
\small
\begin{tabular}{|p{0.9\columnwidth}|}
    \hline
    \textbf{Output of linearization for the table in Figure~\ref{fig:sample}:} \\
    ~\\
    ... ELP at Year~1998 is~2.465. ELP at Year~1999 is~2.476. ELP at Year~2000 is~2.504. ELP at Year~2001 is~2.490. ELP at Year~2002 is~2.482. ELP at Year~2003 is~2.473. \\
    \hline
\end{tabular}
\caption{Example output of Linearization.}
\label{table:linearization}
\end{table}

Firstly, we compared with three table-to-text generators that achieved state-of-the-art results on the recent LogicNLG dataset~\cite{Chen2020} which, similar to our \DATASETNAME, requires synthesizing data in multiple cells. These generators are open source. \textbf{Field-Infusing} employs LSTM to encode each table into a sequence of vectors and then applies Transformer to generate text. \textbf{GPT-Linearization} linearizes each table as a paragraph by horizontally scanning the table and concatenating each content cell with its row header and column header into a sentence. Table~\ref{table:linearization} illustrates such a paragraph. The resulting paragraph is then fed into GPT-2 to generate a new text. \textbf{Coarse-to-Fine} is an enhanced version of GPT-Linearization. It adopts a two-step text generation process: generating a template and then filling it.

Furthermore, we implemented an enhanced version of GPT-Linearization and Coarse-to-Fine, referred to as \textbf{GPT-Linearization$^+$} and \textbf{Coarse-to-Fine$^+$}, respectively. At the beginning of the paragraph fed into GPT-2, we inserted the scenario passage and question to enable GPT-2 to perform question-aware text generation.

All the above supervised table-to-text generators were trained based on sentences with positive usefulness labels.

\subsubsection{Unsupervised Methods.}

We also compared with two naive table-to-text generators.

Recall that GPT-Linearization generates a paragraph from tables and then feeds it into GPT-2 to generate a new text. We implemented \textbf{Linearization}. It directly outputs the generated paragraph without feeding it into GPT-2.

\begin{table}[t!]
\centering
\small
\begin{tabular}{|p{0.9\columnwidth}|}
    \hline
    \textbf{Output of templation for the table in Figure~\ref{fig:sample}:} \\
    ~\\
    ... ELP at Year~2000 is~2.504. ... ELP decreases between Year~2000 and~2003. ... ELP generally increases and then decreases. ... ELP reaches a maximum of~2.504 at Year~2000. ... ELP is relatively large between Year~2000 and~2002. ... \\
    \hline
\end{tabular}
\caption{Example output of Templation.}
\label{table:templation}
\end{table}

Besides, we implemented \textbf{Templation}. It generates a paragraph consisting of all the sentences~$\hat{S}$ generated by our templates. Sentences are sorted in ascending order of length so that if the paragraph has to be truncated by the maximum sequence length of K-BERT, the largest number of sentences can be retained. Table~\ref{table:templation} illustrates such a paragraph.

\subsubsection{Gold-Standard Sentence.}

Last but not least, we used manually annotated gold-standard sentence as a reference.

\subsection{Implementation Details}
\label{sec:exp-impl}

We performed 5-fold cross-validation. For each fold, we split \DATASETNAME\ into 80\%~for training and 20\%~for test. For model selection, we relied on an inner holdout 80\%/20\% training/development split. We ran all the experiments on TITAN RTX GPUs.

For K-BERT, we used BERT-wwm-ext~\cite{chinese-bert-wwm}, a pre-trained Chinese language model as the underlying language model. We set $\text{maximum sequence length}=256$, $\text{self-attention layer}=12$, $\text{hidden units}=768$, $\text{epochs}=15$ for MRC and template-level ranking, $\text{epochs}=5$ for  sentence-level ranking, $\text{batch size}=8$ for MRC, $\text{batch size}=16$ for template-level ranking and sentence-level ranking, $\text{learning rate}=1e\text{--}5$, and $\text{attention heads}=12$. For knowledge base retrieval we set $\epsilon=2$. Inspired by \citet{DBLPconf/aaai/JinGKCH20}, for the K-BERT model in our MRC method (but not the one in \MODELNAME), we coarse-tuned it on $\text{C}^3$~\cite{sun2019investigating}, a Chinese MRC dataset.


For GPT-2, we used $\text{CDialGPT2}_\text{LCCC-base}$~\cite{wang2020chinese}, a pre-trained Chinese GPT-2 model. For $\text{CDialGPT2}_\text{LCCC-base}$, and for LSTM and Transformer in Field-Infusing, we followed the recommended hyperparameter settings in their original implementation.

For our \MODELNAME, by default we set $k=2$ to only select the top-2 generated sentences for MRC. We will report a comparison in different settings of~$k$.

\subsection{Evaluation Metrics}

To evaluate TSQA, we measured \textbf{accuracy}, i.e.,~the proportion of correctly answered questions.

To evaluate sentence ranking, we measured the quality of the whole ranked list of all the sentences~$\hat{S}$ generated by our templates. We used two standard information retrieval evaluation metrics: Mean Average Precision~(\textbf{MAP}) and Mean Reciprocal Rank~(\textbf{MRR}).

\section{Experiment Results}
\label{sec:exp-result}

We report average results on the test sets over all the folds.

\subsection{Results on TSQA}

\subsubsection{Comparison with Baselines.}

\begin{table}[t!]
    \centering
    \small
    \begin{tabular}{|l|l|}
        \hline
        & Accuracy \\
        \hline
        Field-Infusing & 0.353~$^\bullet$ \\
        GPT-Linearization & 0.370 \\
        Coarse-to-Fine & 0.367 \\
        GPT-Linearization$^+$ & 0.348~$^\bullet$ \\
        Coarse-to-Fine$^+$ & 0.359~$^\circ$ \\
        Linearization & 0.235~$^\bullet$ \\
        Templation & 0.243~$^\bullet$ \\
        \MODELNAME & \textbf{0.397} \\
        \hline
        Gold-Standard Sentence & 0.418 \\
        \hline
    \end{tabular}
    \caption{Accuracy of TSQA. We mark the results of baselines that are significantly lower than \MODELNAME\ under $p<0.01$~$(^\bullet)$ or $p<0.05$~$(^\circ)$.}
    \label{table:main_results}
\end{table}

Table~\ref{table:main_results} shows the accuracy of TSQA achieved by each method. Our \MODELNAME\ outperforms all the baselines by 2.7--16.2~percent of accuracy.

\MODELNAME\ exceeds three state-of-the-art table-to-text generators, i.e.,~Field-Infusing, GPT-Linearization, and Coarse-to-Fine, by 2.7--4.4~percent of accuracy.

The enhanced version of these generators that we implemented, i.e.,~GPT-Linearization$^+$ and Coarse-to-Fine$^+$, exhibit surprisingly worse performance than their original version. Their generation methods are significantly inferior to our \MODELNAME\ by 3.8--5.1~percent of accuracy.

The two naive generators, i.e.,~Linearization and Templation, produce much noise information for MRC and achieve accuracy even lower than random guess~(i.e.,~0.25). It demonstrates the necessity of ranking and selecting generated sentences.

The accuracy of using gold-standard sentence is~0.418. On the one hand, compared with the accuracy~0.397 of our \MODELNAME, it suggests that there is still room for improving our templates and/or our sentence ranking model. On the other hand, the achieved accuracy is not satisfying. To improve the overall performance of our approach, we need to combine our \MODELNAME\ with novel MRC methods that are more powerful than K-BERT to meet the unique challenges raised by the \DATASETNAME\ dataset. This will be our future work.

\subsubsection{Varying~$k$.}

\begin{table}[t!]
    \centering
    \small
    \begin{tabular}{|l|c|c|c|c|c|c|}
        \hline
                & $k=1$ & $k=2$ & $k=3$ & $k=4$ & $k=5$  \\
        \hline
        Accuracy & 0.390 & 0.397 & 0.352 & 0.343 & 0.330 \\
        \hline
    \end{tabular}
    \caption{Accuracy of TSQA by varying~$k$ in \MODELNAME.}
    \label{table:changek}
\end{table}

Table~\ref{table:changek} shows the accuracy of TSQA achieved by our approach under different settings of~$k$. Increasing~$k$ from~1 to~2 (the default value), the accuracy remains stable. Further increasing~$k$ to~3 or larger, the accuracy drops substantially, probably influenced by the extra noise information. It is thus important to rank generated sentences and only select those useful for answering the question.

\subsubsection{Ablation Study.}

\begin{table}[t!]
    \centering
    \small
    \begin{tabular}{|l|l|l|}
        \hline
        & Accuracy\\
        \hline
        \MODELNAME & \textbf{0.397} \\
        \MODELNAME\ w/o tabular data & 0.372 \\
        \MODELNAME\ w/o domain knowledge & 0.380 \\
        \hline
    \end{tabular}
    \caption{Accuracy of TSQA (ablation study).}
    \label{table:tsqaablation}
\end{table}

To analyze the usefulness of tabular data and domain knowledge in TSQA, we implemented two variants of our approach. The first variant ignored tabular data. The second variant ignored domain knowledge.

Table~\ref{table:tsqaablation} shows the accuracy of TSQA achieved by each variant. Compared with the full version of our approach, the accuracy of both variants decrease, by 2.5~percent of accuracy without tabular data and by 1.7~percent of accuracy without domain knowledge. The results reveal the usefulness of tabular data and of domain knowledge.

\subsection{Results on Sentence Ranking}

We compared our sentence ranking model with a strong baseline method:
\textbf{RE2}~\cite{yang-etal-2019-simple}. This state-of-the-art text matcher is open source. We employed it to compute the semantic relevance of each generated sentence in~$\hat{S}$ to the question.
Specifically, we used RE2 as a text pair classifier to predict a ranking score for each generated sentence conditioned on (i.e.,~paired with) a concatenation of the scenario passage and question. We followed the recommended hyperparameter setting in its original implementation.

Table~\ref{table:ranking} shows the quality of sentence ranking computed by each method. Our \MODELNAME\ exceeds RE2 by 5.2~percent of MAP and by 6.0~percent of MRR. Paired t-tests show that all these differences are statistically significant under $p<0.01$.






\subsection{Error Analysis}

We randomly sampled 100~questions to which our approach provided incorrect answers. We analyzed the question answering process and identified the following three main causes of errors. Multiple causes could apply to a question.

\subsubsection{Knowledge Base.}
For 76\%~of the errors, there is a lack of necessary domain or commonsense knowledge for answering the question, such as the location of a particular lake. It suggests expanding our knowledge base. However, this is orthogonal to our technical contribution.

\subsubsection{Reasoning Capabilities.}
For 62\%~of the errors, more advanced reasoning skills are needed. For example, some questions require multi-hop math calculations over a group of related domain concepts. K-BERT as a language model cannot calculate. It is also impracticable to encapsulate such extremely complex operations with predefined templates. Therefore, it suggests incorporating specific calculators and powerful reasoners into MRC models.

\subsubsection{Sentence Ranking.}
For 54\%~of the errors, our sentence ranking model chooses a sentence that is not useful for answering the question. Indeed, some templates and their generated sentences are linguistically similar though logically different, e.g.,~\emph{is relatively large}, \emph{reaches maximum}, and \emph{increases}. This sometimes challenges our sentence ranking model as well as our MRC method. We will focus on this problem in the future work.

\begin{table}[t!]
    \centering
    \small
    \begin{tabular}{|l|l|l|}
        \hline
        & MAP & MRR \\
        \hline
        RE2 & 0.434~$^\bullet$ & 0.461~$^\bullet$ \\
        \MODELNAME & \textbf{0.486} & \textbf{0.521} \\
        \hline
    \end{tabular}
    \caption{Quality of sentence ranking. We mark the results of baselines that are significantly lower than \MODELNAME\ under $p<0.01$~$(^\bullet)$.}
    \label{table:ranking}
\end{table}

\section{Conclusion}
\label{sec:conclusion}

Our study aims at solving TSQA in a fully automated manner to avoid manually interpreting tabular data using natural language descriptions as done in previous research. To support this study, we constructed and published the first dataset \DATASETNAME\ that is dedicated to the TSQA task. With only six templates encapsulating predefined operations for synthesizing tabular data in various ways, we covered most needs about tables in \DATASETNAME\ but then, the problem turned into selecting, among a large number of sentences generated from templates, the most useful ones for answering the question. Our proposed model effectively integrates sentence-level and template-level ranking, and exploits the scenario passage, question, and domain knowledge by fusing their information with K-BERT. Our approach has the potential to be adapted to other AI applications that require table comprehension and explanation.

Although our approach outperformed a variety of strong baselines in the experiments, its accuracy is still not satisfying. Following the results of our error analysis, for the future work, we plan to enhance our sentence ranking model with more powerful semantic matching techniques. We will also extend our MRC method to perform math calculation and logical reasoning over an expanded knowledge base.

\section{Acknowledgments}
This work was supported by the National Key R\&D Program of China (2018YFB1005100). We thank the annotators for their efforts and thank the reviewers for their comments.

\bibliography{main.bib}

\end{document}